\RequirePackage{fix-cm}
\documentclass[smallextended]{svjour3}       
\smartqed  
\usepackage{graphicx}
\usepackage{booktabs}
\usepackage{tabularx}
\usepackage{hyperref}
\usepackage{soul}

\usepackage[table,xcdraw]{xcolor}
\usepackage{epstopdf}

\usepackage{xstring}
\usepackage{subcaption}
\usepackage[table]{xcolor}
\journalname{Language Resources and Evaluation}

\newcommand{\vbl}{{\vrule width 1.1pt}}
\newcommand{\hbl}{\noalign{
\hrule height 1.1pt
}}
\newcommand{\pad}{1.3}

\begin{document}

\title{Learnings from Technological Interventions in a Low Resource Language
}
%

\subtitle{Enhancing Information Access in Gondi}


\author{Devansh Mehta$^{**}$\thanks{** Equal Contribution}
\and
Harshita Diddee$^{*}$\footnotemark[1]
\and
Ananya Saxena
\and
Anurag Shukla
\and
Sebastin Santy
\and
Ramaravind Kommiya Mothilal
\and
Brij Mohan Lal Srivastava
\and
Alok Sharma
\and
Vishnu Prasad
\and
Venkanna U
\and
Kalika Bali
}


\institute{Devansh Mehta \at
            Voicedeck Technologies \\
              \email{devansh76@gmail.com}
          \and
          Kalika Bali \at
          Microsoft Research India \\
          \email{kalikab@microsoft.com}
}

\date{Received: date / Accepted: date}

\maketitle

\begin{abstract}
The primary obstacle to developing  technologies for low-resource languages is the lack of representative, usable data. In this paper, we report the deployment of technology-driven data collection methods for creating a corpus of more than 60,000 translations from Hindi to Gondi, a low-resource vulnerable language spoken by around 2.3 million tribal people in south and central India. During this process, we help expand information access in Gondi across 2 different dimensions (a) The creation of \textbf{linguistic resources} that can be used by the community, such as a dictionary, children's stories, Gondi translations from multiple sources and an Interactive Voice Response (IVR) based mass awareness platform; (b) Enabling its use in the digital domain by developing a \textbf{Hindi-Gondi machine translation model}, which is compressed by nearly 4 times to enable it's edge deployment on low-resource edge devices and in areas of little to no internet connectivity. We also present preliminary evaluations of utilizing the developed machine translation model to provide assistance to volunteers who are involved in collecting more data for the target language. Through these interventions, we not only created a refined and evaluated corpus of 26,240 Hindi-Gondi translations that was used for building the translation model but also engaged nearly 850 community members who can help take Gondi onto the internet. 

\end{abstract}

\section{Introduction}
\label{intro}

In the present era of globalization and integration of technology into almost every aspect of life, native speakers are turning to dominant languages at a faster rate than ever before. As seen in Figure\ref{intro}, around 40\% of all languages in the world face the danger of extinction in the near future \cite{31}, while 95\% have lost the capacity to ascend to the digital realm \cite{kornai2013digital}. When a language spoken in a particular community dies out, we lose a vital part of the culture that is necessary to completely understand it. At a bare minimum, languages need to be integrated with the Internet to give them a fighting chance at survival.

In this work, we design, deploy and critically assess technological interventions that aim to create a digital ecosystem of Gondi content, a South-Central Dravidian tribal language spoken by the Gond tribe in Central India. The case of Gondi is representative of many languages across the world and presents a unique case study of how a language can be in danger despite having all the ingredients of sustainability such as (1) long historical continuity (2) a population of 3 million people speaking it and (3) widely spoken in around 6 states of India with various dialects and forms. The complexities arise as Gondi is a predominantly spoken language with no single standard variety but a number of dialects, some mutually unintelligible \cite{beine1994sociolinguistic}. Our objectives with these interventions is creating a repository of linguistic resources in Gondi that can be used for: (1) building language technologies like machine translation or speech to text systems that are essential for taking Gondi onto the internet; and (2) expanding the information available to the Gond community in their language.


There needs to be groundwork and identification of the real problems that can be solved by deployment of language technologies in minority communities \cite{dearden2015ethical}, as there are legitimate technological and ethical concerns surrounding the use of technology in low-resource languages \cite{joshi2019unsung,10.1145/3530190.3534792}. Technical systems aren't simply ethically neutral exchanges of information and those that work in isolation from social contexts can be detrimental to the ecosystem of minority language communities. At the same time, inaction on the part of language technologists is an ethically dubious proposition in its own right, as minority communities are currently forced to learn mainstream languages for availing the benefits of the Internet and to access power structures enabling socio-economic mobility. We believe it to be fairly non-controversial to maintain that anyone should be able to surf the internet in the language that they are comfortable in. Ethical issues only arise on the methods one uses to reach this goal.

\begin{figure}[!t]
    \centering
    \includegraphics[width=0.8\linewidth,keepaspectratio]{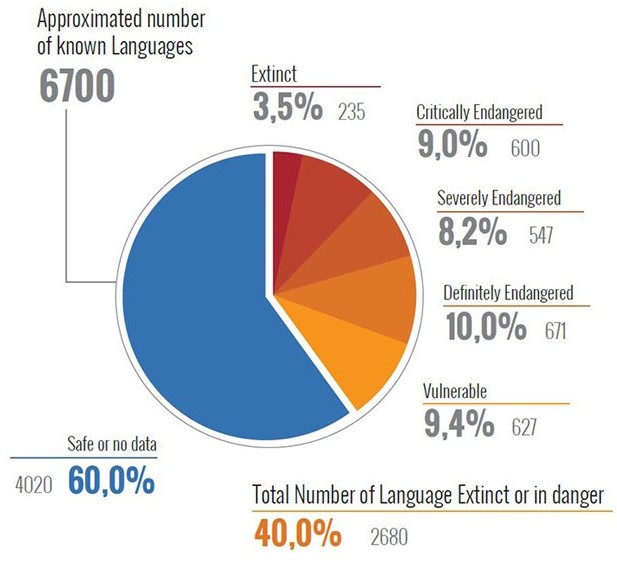}
    \caption{UNESCO 2017 World Atlas: Demonstrating the endangered state of the languages of the world.}
\end{figure}






Our focus when working with the Gond community is centered around devising novel approaches for creation, evaluation and dissemination of Gondi content, unlike well-resourced languages where the focus is more on engineering language technologies using abundantly available data. As far as possible, we have ensured that the technology interventions described in this study were led by the Gond community and non-profit organizations working with them.


The first intervention is creating a 3,500 word Gondi dictionary that is accessible to the community as an Android app. The second is 230 children's books that were translated by the Gond community in a 10 day workshop. The third is crowdsourcing translations from community volunteers via an Android app, the WhatsApp Business API and an assistive translation tool. The fourth intervention is a phone number that community members could call to gain awareness about a local election in their area, upon completion of which they earn mobile credits. These interventions respectively resulted in facilitating 80-100 community members to create a 3,500 word dictionary; 20 community members that translated 230 children's stories from Hindi to Gondi; 219 community members that translated nearly 55,000 sentences; and 557 native speakers of Gondi that learned about elections in their own language and who can be called for future workshops. 

This data was used to develop a Hindi-Gondi Machine Translation Model\footnote{https://github.com/microsoft/INMT-lite}, compressed for streamlined operation on low-resource edge devices for use by the community. The model was evaluated by community members that engaged with us through these interventions and got an average Direct Assessment score \cite{specia-etal-2020-findings-wmt} of 63, indicating the semantic accuracy of our model's output. We also run preliminary experiments for evaluating the model's ability in assisting community members with translations. Our results show promising efficacy with more than 50\% of the suggestions being accepted.


\section{Context}

\begin{figure}[!t]
    \centering
    \fcolorbox{lightgray}{white}{\includegraphics[width=0.8\linewidth,keepaspectratio]{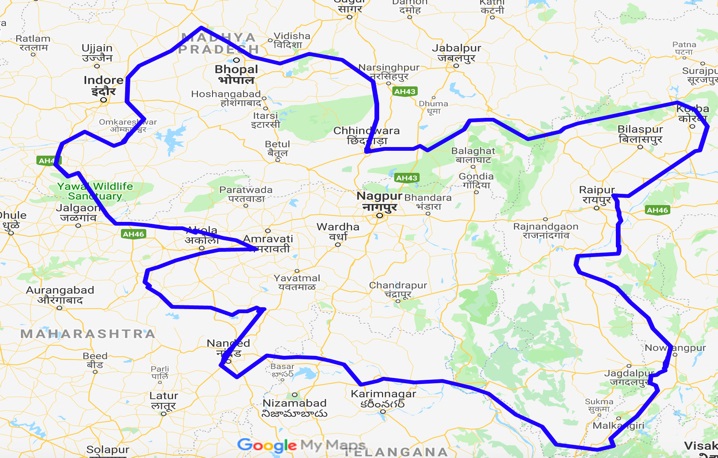}}
    \caption{Gondi speaking areas in India}
    \label{fig:map}
\end{figure}

According to the 2011 census \cite{chandramouli2011census}, the total population of the Gond tribe is approximately 11.3 million. However, the total Gondi speaking population is only around 2.7 million. That is, only about 25 percent of the entire tribe now speaks it as a first language. UNESCO's Atlas of the Worlds Languages in Danger \cite{moseley2010atlas} lists Gondi as belonging to the vulnerable category. There is an added difficulty of creating resources for Gondi due to the linguistic heterogeneity within the Gond community. As seen in Figure\ref{fig:map}, Gondi is spread over 6 states in India. It is heavily influenced by the dominant language of each state to the point where a Gond Adivasi from Telangana (a Telugu speaking Southern state), finds it difficult to understand a Gond Adivasi from Madhya Pradesh (a Central state with Hindi as the dominant language). Recent scholarship has found the influence of these dominant languages to be a major factor in language loss in the Gond community \cite{boruah2020language}.

As a predominantly oral language, the proportion of Gondi speakers is expected to go down further as opportunities are shrinking to hear the language spoken outside of their everyday surroundings. All India Radio, the only radio station in India allowed to broadcast news, does not have regular Gondi news bulletins. There is no TV station or channel catering to Gondi speakers. There is also a severe dearth of online content in Gondi, resulting in members of the tribe having to learn a mainstream language to enjoy the benefits of internet connectivity. ``The Wire"\footnote{Gondi Bulletin: \url{https://youtu.be/M3q2ycJ\_U7g}} is one of the few Indian news outlet that publishes a news bulletin in Gondi. However, the Gondi spoken in their broadcast caters to the Gond Adivasis from the Central states of India and it is difficult for Gonds from the Southern states to understand the content.

Gondi is also not included in the 8\textsuperscript{th} Schedule of the Indian Constitution, with the result that education and exams for government jobs cannot be administered in the language. The deleterious effects of the marginalization of Gondi on their community are manifold. Gondi is considered the lingua franca of the local insurgents, who use their knowledge of the language and the perceived neglect by the government to recruit candidates from the tribe to join their civil war against the state \cite{kumar2019gondi}. Further, there are high dropout rates among monolingual children that speak Gondi as a first language. A UNESCO study found that children whose mother tongue is not the medium of instruction at primary school are more likely to fail in early grades or drop out \cite{buhmann2008mother}, which in turn increases the chances of them joining the insurgency. Working with the Gond tribes on reviving their language is thus important not just for cultural reasons, but may also serve as an instrument for bringing peace to their society. These factors can be mitigated provided there is political will and a standardized version of the language that is agreed upon. While the younger, bilingual generation in the Gond tribe attach less importance to their mother tongue, it was not difficult to find community members worried about the extinction of their language and it proved easy to recruit participants for the workshops we held, even without any financial incentive.




\section{Related Work}

The first section focuses on how we address the constraints of efficiently collecting data through traditional channels for low-resource languages. The second section situates our study within existing language preservation and revitalization programs and lays out our unique contributions to this prior literature.

\paragraph{\textbf{Data Collection Channels}}
\label{sec:data-collection-channels}

Traditional channels of data collection can vary across 3 dimensions (a) Channel of Data Collection (b) Quality of Collected Data and (c) Quantity of Collected Data. There are two common channels of data collections: automated mining and human reliant crowdsourcing. Mining includes a wide spectrum of methods varying from crawling data off the web to utilizing content after digitization by systems such as Optical Character Recognition. Unfortunately, these channels assume the existence of  data resources in a specific form: for example, Automatic or Semi Automatic methods like Web mining expects the data to be digitally readable, while OCR expects data to be digitally consumable by an existing service, which might not be the case for many languages, especially low-resource ones \cite{rijhwani2020ocr}. 

In our study, we rely on crowdsourcing large quantities of data through human reliant channels. This approach is fraught with logistical challenges around participant recruitment \cite{nangia2021ingredients}, the cost of response validation, and mitigating subjectivity in responses \cite{goldman-etal-2018-strategies}, as well as ethical considerations pertaining to devising  fair pay \cite{kummerfeld-2021-quantifying} \cite{shmueli2021beyond} and maintaining response diversity by controlling for human bias in the inputs provided by a crowdsourced audience \cite{liu-etal-2022-toward}. Our study adds novel insights to this prior literature by building a dataset from community volunteers and staff at partner organizations without the provision of any fiscal incentive, which is often a given attribute of human reliant data crowdsourcing channels. We find that that a volunteer led workforce for data generation can yield higher quality datasets by self-selecting for those motivated to preserve their language.

Prior work has identified quality constraints for corpora arising out of human-collection channels \cite{10.1162/tacl_a_00447} \cite{caswell-etal-2020-language}. Specifically, noise derivative of code-mixing, non-linguistic characters, short/meta-data segments, untranslated/copied content has been shown to considerably harm the output of neural machine translation (NMT) systems \cite{khayrallah2018impact}. For high resource languages, the sheer quantity of a corpus is usually significant enough to allow the selection of a subset of high-quality sentences from the larger set using intelligent validation techniques, especially since some percentage of noisy data is shown to improve generalization robustness of NMT models. Training and evaluation corpora in low-resource languages may not be as effective due to the paucity of data \cite{10.1162/tacl_a_00447}. We observe this explicitly with our interventions, since a large proportion of our data (nearly 60K data samples that we collected) were sifted to a set of ~27K samples, greatly reducing the effective yield of our collection channels. Our work thus presents a case study on developing 
alternative methods of data collection and evaluation for low-resource languages.




\paragraph{\textbf{Language Technologies for Revitalization and Preservation}}
\label{sub:language-preservation}

Preservation programs generally attend to the need to develop data corpora in that language \cite{abraham2020crowdsourcing}, \cite{article}, \cite{mus2021toward}. However, the notion that language documentation or artifact creation can independently bring about revitalization is largely  dismissed by the community  \cite{zariquiey-etal-2022-cld2}. Community-centric technical interventions that can infuse the language technology into the community for sustained use are increasingly coming under focus. One dimension of this focus has been towards developing fundamental NLP tools for low-resource languages such as Part-Of-Speech taggers \cite{finn-etal-2022-developing}, parsers \cite{bowers-etal-2017-morphological}, etc. Another dimension is the development of more end-user oriented language technologies such as edge-deployments \cite{bettinson-bird-2017-developing} \cite{hermes2013ojibwe} \cite{bird2014aikuma} and interactive interfaces that can be consumed rapidly by the community \cite{little-2017-connecting} \cite{adams-etal-2021-user}. A critical point to note here is that access bottlenecks can limit the consumption of outcomes from revitalization and preservation endeavours. For instance, Roche highlights the political nature of language endangerment, pointing out that low-resource language communities are usually under-resourced in other aspects of life \cite{roche2020abandoning}. Leibling et al \cite{10.1145/3313831.3376261} discuss these constraints across the adaptation of translation systems to edge devices across three globally distributed and divergent demographic groups. We contribute to this identified gap by creating a software artifact that displays NMT translations on low cost devices like smartphones, which can be used by other speech communities and researchers working with marginalized populations.

 Due to the well-studied constraints of data collection through traditional channels \cite{elazar-etal-2020-extraordinary} \cite{millour-fort-2020-text} \cite{liu-etal-2022-toward}, our work explores newer mechanisms for collecting low-resource language data that are of wider relevance. For example, our Adivasi Radio translation app allowed users to bulk download a corpus of sentences for translation, which they could then complete even from areas without Internet. Another area our work sheds new light on is reducing the cognitive burden faced by communities in understanding an entirely novel interface for data provision, as discussed in prior work like \cite{10.1145/2858036.2858448} \cite{10.1145/2556288.2557155}. Our study uses low friction approaches for collecting translations such as through the WhatsApp Business API, which is well integrated into many communities. Lastly, our work briefly touches upon the attempt to integrate assistive translation interfaces to provide sub-optimal help in data collection efforts \cite{10.1145/3290605.3300461} \cite{10.1145/3530190.3534792}. 
 
Overall, this paper is a case study in collaborating with target communities for creating linguistic resources to enable the development of language technologies that can then give rise to a digital ecosystem for such languages. In this regard, we follow the lead set by Kornai and Bhattacharyya in building online tools for low-resource languages like spellchecks that can be incrementally refined over time \cite{kornai2014indian}. Our work distinguishes itself by empirically studying methods of community engagement that are cognizant of the constraints faced in low-resource environments.

\section{Technological Interventions}

The larger framework of our interventions is that language resource creation feeds into building language technologies and enhancing access to information in that language. We rolled out a series of interventions that created linguistic resources useful to the community, such as children's stories, which in turn fed into building language technologies such as a Hindi-Gondi translation system. These initiatives helped increase information access in Gondi while supporting the end goal of taking Gondi onto the Internet.



\subsection{Gondi Dictionary Development}

This section discusses the need and complexity of building a common dictionary across the various Gondi dialects.

\subsubsection{Motivation} Several researchers have found that mutual intelligibility of Gondi decreases with distance, in part due to the influence of dominant state languages creeping into the various dialects \cite{beine1994sociolinguistic,shapiro1981language,tyler1969koya}. Beine (1994) conducted mutual intelligibility tests across the Gondi speaking areas and found there to be 7 mutually unintelligible dialects of the Gondi language (pp 89). He thus recommended the creation of dialect centres to cover each Gondi speaking region, with literacy materials separately developed for each center. These workshops helped determine whether the community wanted separate efforts for each dialect or a common effort towards one language.

\subsubsection{Intervention} Prior to the involvement of our research team, a citizen journalism platform called CGNet Swara 
 held 7 workshops beginning from 2014 to develop a Gondi thesaurus containing all the different words used by native speakers from 6 states. Some words, such as water, had as many as 8 different words for it. At the 8\textsuperscript{th} workshop in 2018, which saw more than 80 people in attendance, the thesaurus was developed into a dictionary containing 3,500 words.

\begin{figure}[!t]
    \centering
        \begin{subfigure}{0.49\linewidth}
            \fcolorbox{lightgray}{white}{\includegraphics[width=\columnwidth,keepaspectratio]{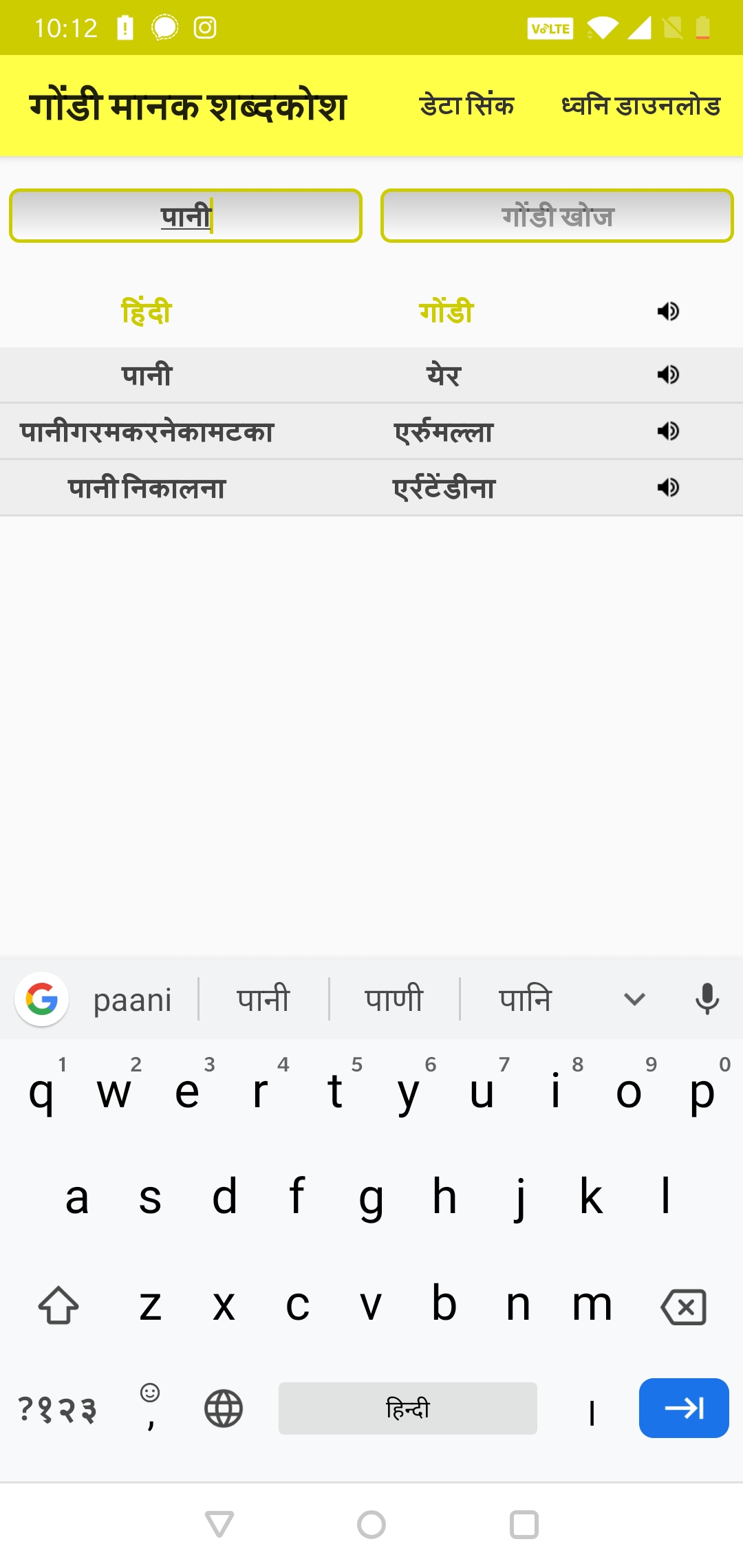}}
        \end{subfigure}\hfill
    \caption{Gondi Manak Shabdkosh, a dictionary of 3,500 words}
    \label{fig:gondimanak}
\end{figure}

The dictionary was made into an Android app, Gondi Manak Shabdkosh\footnote{\url{aka.ms/gondi-dictionary}} \footnote{\url{aka.ms/indian-express-gondi-dictionary}}, depicted in Figure\ref{fig:gondimanak}. It allows users to enter a Hindi or Gondi word and hear or read its equivalent translation, similar to the Ma! Iwaidja dictionary app\footnote{\url{ma-iwaidja-dictionary.soft112.com/}} without the wheel based interface for conjugation and sentence formation.

\subsubsection{Takeaways} CGNet Swara reported an overwhelming consensus by the community to remain as one language. Gondi is primarily used within its local speech community and participants recognized the need to develop a common vocabulary and spelling for its emergence as a standard language. The three techniques commonly used in standardizing languages are comparative (linguistic reconstruction to build a mother tongue for all), archaizing (deriving a variety from older, written texts) and the statistical (combining the different dialects having the widest usage) \cite{haugen1966dialect}, with most participants favoring the latter approach.

CGNet's initial aim with the dictionary app was studying whether it can allow some basic communication and learning to take place in primary schools where a teacher does not know Gondi while their student is a monolingual speaker. However, our approach of finding a central dialect mediating between the extremes resulted in a variety of Gondi promoted as everyone's language that became nobody's language. For example, monolingual Gonds may only be aware of their local word for water, but if it hasn't been selected in the dictionary then its use in such scenarios would be limited. In retrospect, a thesaurus might have been more applicable for this particular use case. The dictionary provided greater utility in pretraining our translation model. Following Wang et al \cite{wang2022expand} we augmented digitally consumable monolingual data to improve performance of the model, as demonstrated in Table \ref{tab:model-scores}. 



{}

\subsection{Creating children's books in Gondi through Pratham Books}

We narrate our efforts at translating children's stories into Gondi for use in schools and as a data source for language technologies.

\subsubsection{Motivation} Imparting basic, primary education in a child's mother tongue inculcates pride in their language and cultural identity formation among younger generations. It also targets the most important stakeholder for ensuring continued language use. More practically, having Gondi as a medium of instruction or a subject at schools can help increase enrollment among monolingual Gond tribes within the education system and decrease drop-out rates among indigenous children not conversant in Hindi. This intervention focused on developing the necessary linguistic resources in Gondi that could serve as educational material.



\subsubsection{Intervention} In 2018, our research team conducted a 10 day translation workshop with CGNet Swara and Pratham Books, a nonprofit publisher with the motto "Book in every child's hands". 20 Gond tribals from three states of India came together and translated 230 children's books from Hindi to Gondi, some of which introduced Gondi children to climate change for the first time  \footnote{Article: \url{aka.ms/news18-climatechange}}. These stories were published on the Storyweaver platform \footnote{\url{storyweaver.org.in/}}, an initiative of Pratham Books that hosts more than 15,000 children's stories in various languages and dialects. To the best of our knowledge, this is the first online repository of children's stories in Gondi \footnote{Article: \url{aka.ms/hindu-gondi-nextgen}\footnote{\url{aka.ms/storyweaver-gondi}}}. They were also printed out and distributed in primary schools in the Gondi speaking districts of Chhattisgarh, with efforts now ongoing to convince the state government to include them as part of the school curriculum across the tribal belts of the state.


\subsubsection{Takeaways} Although no payment was provided to workshop participants (besides covering food, travel and lodging), a total of around 20 volunteers translated 8000 sentences over 10 days. Having a volunteer only workforce helped in self-selecting individuals that cared about the languages survival among future generations. We found it most effective to create teams of 4-5 people for translating a story, with more educated participants taking on the role of writing the translation on paper or typing the written translation into the computer interface. Monolingual Gond speakers would help find the right vocabulary for translations, while bilingual speakers would help them understand the Hindi text that needed translating. Having groups complete translations (rather than individuals) resulted in considerable debate around which Gondi words to use for translating a Hindi sentence. This intervention yielded the most high quality data compared to the other efforts, as is evident from our manual evaluation of the data referenced in Table \ref{tab:data-scores}.

\subsection{Crowdsourcing Gondi translations through Adivasi Radio}

Unlike the Pratham Books workshop featuring in-person group translations, Adivasi Radio was geared towards the post pandemic world and allowed volunteers to provide translations from the comfort of their homes.

\subsubsection{Motivation} At the children's book workshop, we found that many participants wanted to continue the translation work from home but there was no avenue for them to do so. Taking inspiration from Aikuma\cite{bird2014aikuma}, we developed Adivasi Radio, an Android application that presents users with Hindi words or sentences for which they need to provide the Gondi translation. In addition to the translation role, Adivasi Radio was designed as the go-to place for native speakers to find outlets and sites publishing Gondi content. 
\begin{figure}[!t]
    \centering
    \begin{subfigure}{0.49\linewidth}
            \fcolorbox{lightgray}{white}{\includegraphics[width=\columnwidth,keepaspectratio]{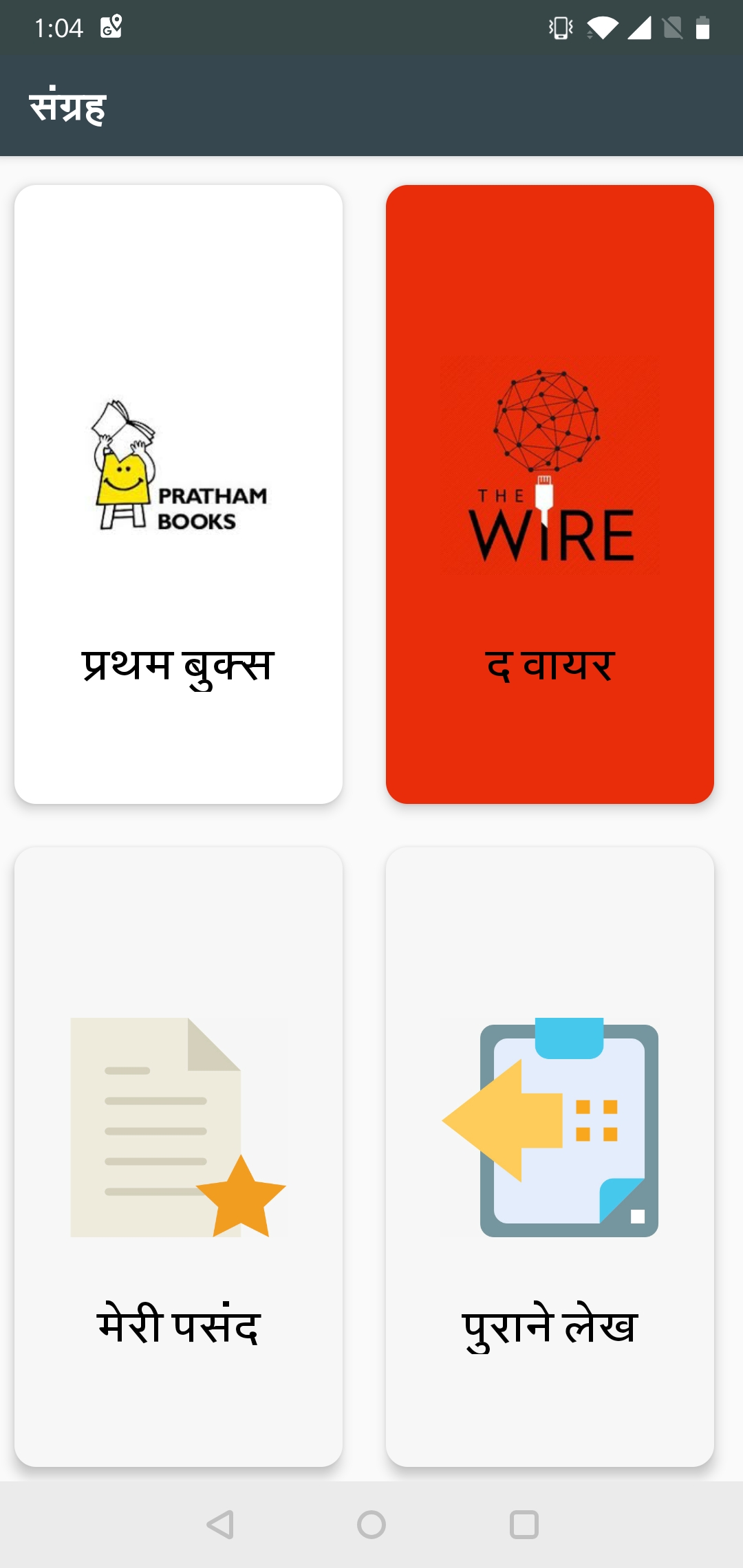}}
        \end{subfigure}\hfill
        \begin{subfigure}{0.49\linewidth}
            \fcolorbox{lightgray}{white}{\includegraphics[width=\columnwidth,keepaspectratio]{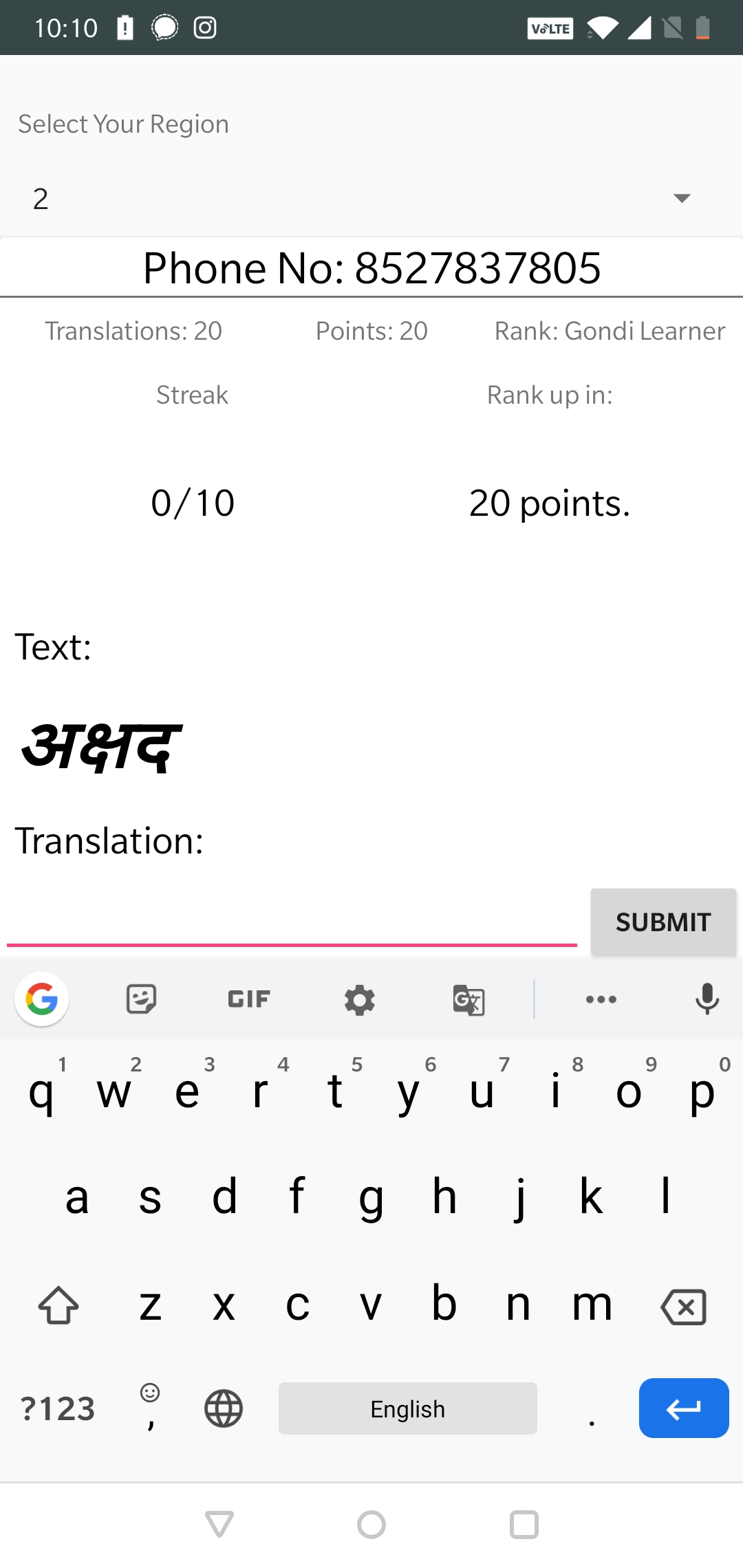}}
        \end{subfigure}\hfill
    \caption{Adivasi Radio app to collect translations and access Gondi content}
    \label{fig:adivasiradio}
\end{figure}{}

\subsubsection{Intervention} 
\begin{figure*}[!t]
    \centering
    \includegraphics[width=\linewidth,keepaspectratio]{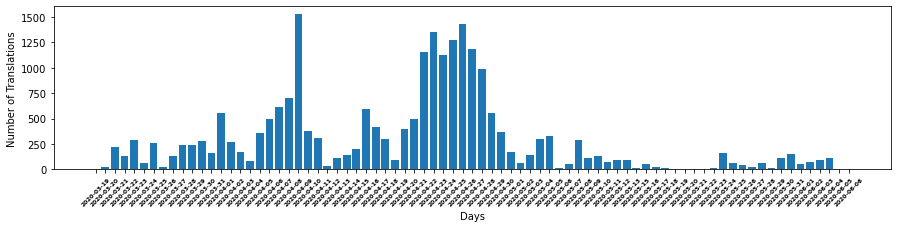}
    \caption{Number of translations collected per day.}
    \label{fig:translation-usage}
\end{figure*}

The app was first deployed between March and June 2020, during the initial months of the coronavirus induced lockdown. The interface of the app can be seen in Figure\ref{fig:adivasiradio} while translations over time are shown in Figure\ref{fig:translation-usage}. In 4 months, 17,756 sentences were translated through the app from 164 unique users. 

6 superusers, most of whom were staff at our partner organization CGNet Swara, accounted for the majority of the translations. Many volunteers gave up after completing a few translations, indicating that sustaining data collection efforts is a challenge, especially without fiscal incentives.

We uploaded the children stories produced with Pratham Books, Gondi stories on CGnet Swara's citizen journalism platform and The Wire's Gondi news bulletin into the app. A Devanagari text-to-speech system read out the written Gondi text so that monolingual speakers could enjoy the content.

\subsubsection{Takeaways} After Adivasi Radio's deployment, we held meetings with the translators to learn about their experience of using the application. We found that several users resided in areas having weak internet connectivity, thus prompting us to update the application so that they could bulk download a large corpus of sentences that they could translate even when they were offline. Similar to the feature used in \cite{d2014mobile,mehta2020facilitating} for enabling citizen journalism in areas without internet, translated sentences would get sent to our server whenever the user managed to connect to the internet. The attention to providing accurate translations was reflected in their request for both a 'skip' feature to avoid providing translations they were unsure of and an option to edit their earlier submissions, as they would sometimes realize mistakes only after completing a translation. However, we found that the skip feature was used to avoid translating long sentences or short paras of three or more sentences. This concern could be alleviated by having them translate sentences forming a cohesive narrative like a Wikipedia article rather than disjointed sentences unrelated to one another. Finally, we did not see significant engagement with the uploaded Gondi content available on the app as most users were bilingual and accustomed to a richer online environment in Hindi.

Our evaluation of the quality of submitted translations reiterated the importance of incorporating rigorous validation checks on data collected through human-channels. Out of the nearly 18000 sentences collected through this channel,  we were only able to utilize 10105 sentences after weeding out transliteration, skipped sentences and gibberish inputs with a high density of special and/or numeric characters.

\subsection{Crowdsourcing Translations through WhatsApp Business API}

A growing body of research advocates integration with existing platforms for coordinating social mobilization instead of creating custom made ones such as an app or website \cite{lambton2020unplatformed,saldivar2019online,starbird2012crowdwork}. The majority of Adivasi Radio users made extensive use of WhatsApp in their daily life, prompting us to explore how we could use this medium to solicit translations from the community. 

\begin{figure}[!t]
    \centering
    \begin{subfigure}{0.49\linewidth}
            \fcolorbox{lightgray}{white}{\includegraphics[width=\columnwidth,keepaspectratio]{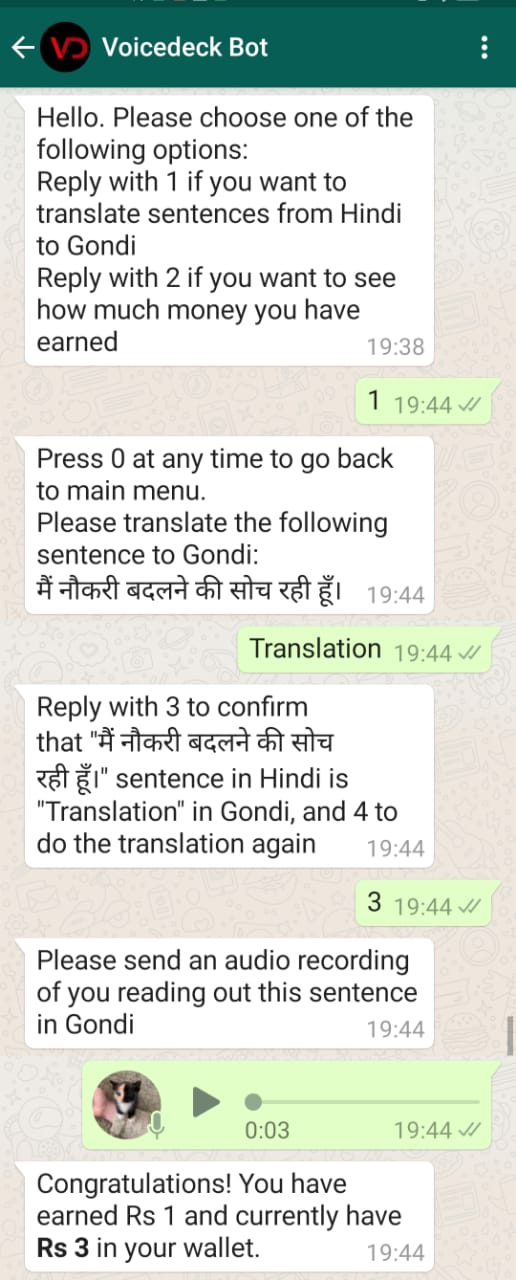}}
        \end{subfigure}\hfill
        \begin{subfigure}{0.49\linewidth}
            \fcolorbox{lightgray}{white}{\includegraphics[width=\columnwidth,keepaspectratio]{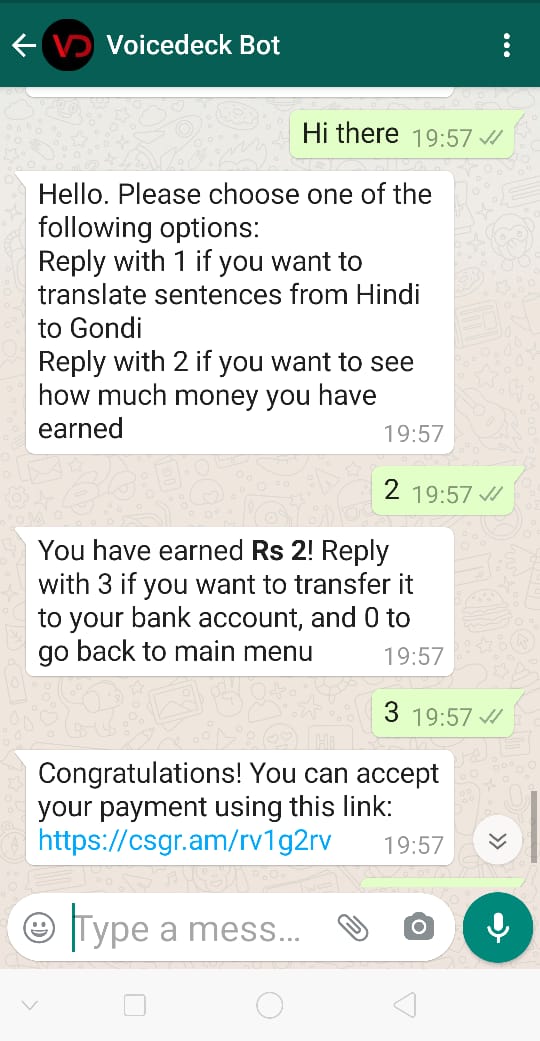}}
        \end{subfigure}\hfill
    \caption{Payments were discontinued for translations received via WhatsApp after finding users were gaming the system}
    \label{fig:whatsapp-api}
\end{figure}{}

\subsubsection{Motivation} During field testing with Adivasi Radio users, we found that some users did not have space on their phones to install our app. In such cases, we had to delete media files and unused apps before we could install the Adivasi Radio application on their phone. Moreover, training had to be provided on how to navigate the interface of our app. These factors limited our ability to reach more members in the community without our physical presence.

We thus applied for - and received - permission to use the WhatsApp Business API, which would enable users to provide translations over WhatsApp directly to our server without any human intervention. We accessed the API through the IMI Connect platform, which has a monthly charge of about USD 350 per month in addition to variable costs per message. 


\subsubsection{Intervention}  To seed usage of the system, we needed users that had a smartphone with regular internet connectivity and were conversant in both Hindi and Gondi. From May 2021 to September 2021, we collected 37,173 sentences from 55 volunteers and staff members at CGNet Swara. 6 superusers accounted for nearly half of the submitted translations. Many translations were written in the latin script (as opposed to Devanagari), resulting in a little less than 12,000 sentences that could be fed into the translation model.

\subsubsection{Takeaways}  

To address declining interest in completing translations from home via Adivasi Radio, we released a probe in February 2021 that for the first time paid users for completing translations via WhatsApp. We found that payment per sentence translated led to deteriorating quality by rewarding speed over quality. Indeed, we saw some of the most devoted community members spend more than 10 minutes on a single sentence, even asking their neighbors for the right word, so that they could get it right. When these members saw their peers provide inaccurate translations and earn more than them, they withdrew their support. We ended up discarding nearly 12,000 sentences collected via this process and switched back to a volunteer driven model. We also observed frustration when the sentence that translators were given timed out and they were provided a new sentence upon logging back in, as they had researched the earlier sentence but there was no option for them to provide it anymore. 

We relaunched the WhatsApp chatbot in June 2021 without the option of earning any money for translations. The majority of the translations came from staff members at CGNet Swara. In the absence of strict oversight and punishments for incorrect translations, we recommend paying a fixed salary to translators rather than one varying by number of sentences translated or other such quantitative metrics. It also proved effective to integrate data collection into the day to day responsibilities of employees at partner organizations.


\subsection{Data Quality}

We requested community members to manually evaluate the data collected and then compare the quality of data collected via each channel.
\subsubsection{Annotation Setup}
\label{sec:annotation-setup}
\begin{figure}[!ht]
\begin{center}
\includegraphics[scale=0.15]{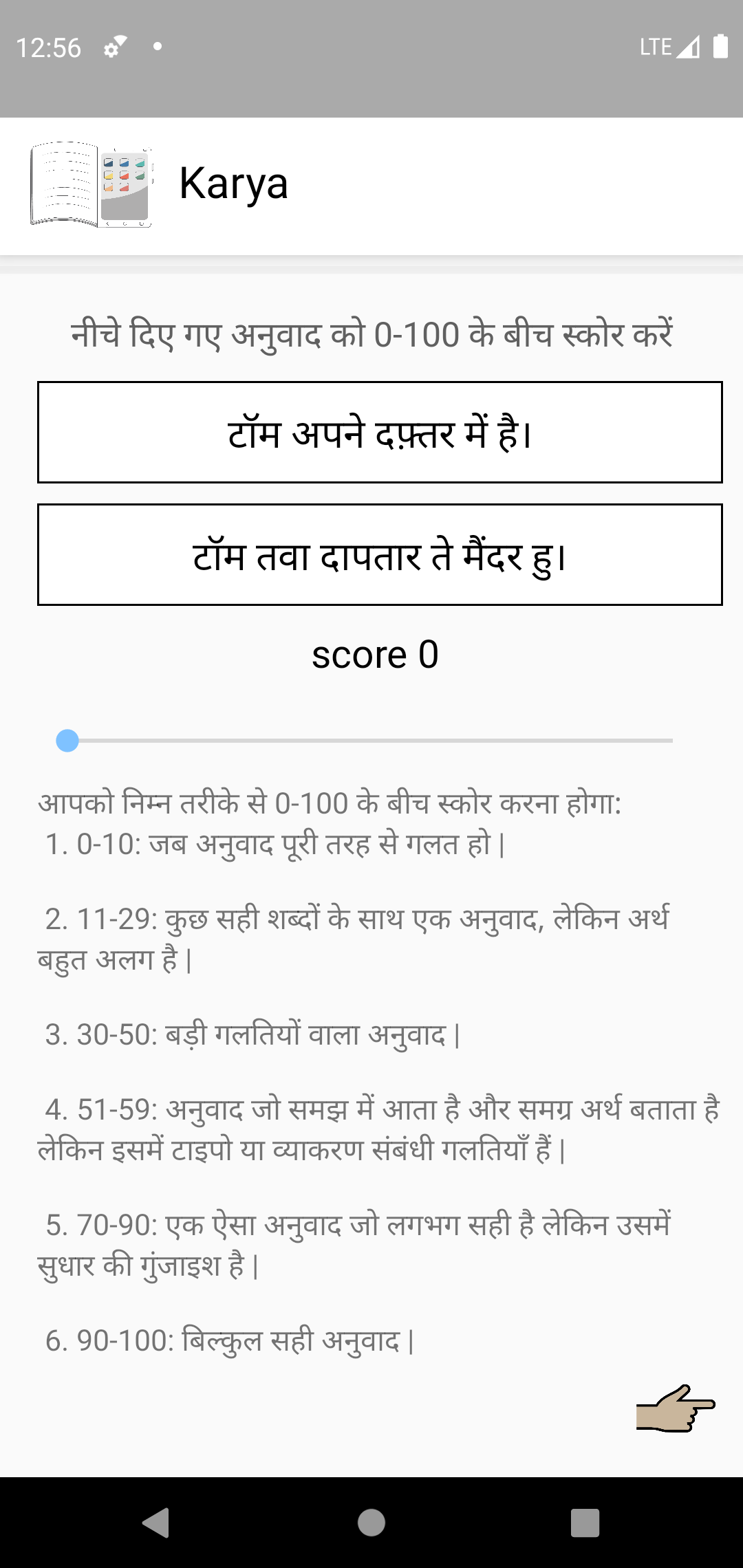} 

\caption{The scoring instructions are specified in Hindi on the annotation screen for quick reference. A score slider is provided for limiting scores to a valid range. }
\label{fig:scoring-interface-karya}
\end{center}
\end{figure}

We recruited 3 Gondi speaking annotators from the Indian states of Maharashtra, Madhya Pradesh and Chhattisgarh to assist us with the manual evaluations of our data and the translation model developed from it. This selection was made bearing in mind the dialectal coverage of our dataset, as most translations were provided by volunteers from these three states. The annotation instructions were made visible to all annotators via the interface we provide to them as shown in Figure\ref{fig:scoring-interface-karya}. We also discussed a set of samples and the mechanism to acquaint them with the scoring system before the evaluation exercise. The participating annotators performed 3 tasks where they scored a provided sample with the Direct Assessment score described in \cite{specia-etal-2020-findings-wmt}. They were compensated using the following system\footnote{ All compensations are specified in Indian National Rupee} - per ranking task, each annotator was given Rs. 5 (Rs. 3 base pay, and Rs. 2 upon completion). Per assistive translation task, they were provided Rs. 5 and per translation task, they were provided Rs.10 ( Rs. 7 base pay, Rs. 3 upon completion ). 

To analyze the quality of the data collected through Pratham Books, Adivasi Radio and WhatsApp API, we requested each annotator to score 60 sentences sampled across any two data sources i.e. 2 annotators score 30 instances per source. These samples were taken from our training corpus, and we apply a sequence length filter in the range of [4, 15] tokens to avoid undue weightage to sentences coming from any one source.

\subsubsection{Results}

We report average direct assessment scores for each data source in Table \ref{tab:data-scores}.
\begingroup
\setlength{\tabcolsep}{3pt} 
\renewcommand{\arraystretch}{1} 
\begin{table}[h]
    \centering
    \small
    \begin{tabular}{lcc}
    \toprule

    \textbf{Channel} &\textbf{Average DA Score} & \textbf{Standard Deviation} \\
    \midrule
     Pratham Books      & 65.4 & 16.1 \\
     Adivasi Radio      & 58.6 & 19.1 \\
     WhatsApp           & 36.5 & 21.9 \\ \hline 
     Model(Hindi-Gondi) & 63.2 & 23.2 \\
    \bottomrule
    \end{tabular}
    \caption{Average Direct Assessment Scores with Standard Deviations for data collected via all channels and the Hindi-Gondi model trained on the data }
    \label{tab:data-scores}
\end{table}
\endgroup

The average DA scores on the sample set indicate that the data collected from Pratham Books is of the best quality, followed by Adivasi Radio and WhatsApp respectively. This could be due to the setup in which participants were placed: synchronous and supervised for Pratham Books and asynchronous and unsupervised for WhatsApp or Adivasi Radio.


As is observed with the standard deviation, the inter-annotator score agreement was the poorest for WhatsApp. The annotators report that WhatsApp data inputs had a very high degree of code-switching, sometimes not even with Hindi but other indic languages. For example, all annotators reported observing Marathi tokens as a part of the provided translations.

\subsection{Machine Translation}

Using the data collected, we train a machine-translation model between Hindi-Gondi and evaluate the efficacy of its output.
\subsubsection{Motivation}

The development of the machine translation model aimed at developing a core-technology that serves the larger goal of improving information access for the community by enabling automated translation of web content from Hindi to Gondi. It also let us evaluate the utility of the data when it is plugged into a real-world NLP usecase.  We envisage the usage of this model in other low-resource languages like that of Bribri, Wixarica and Mundari  \cite{https://doi.org/10.48550/arxiv.2210.15184} to demonstrate the applicability of our artifact across other data-deficient setups. 

\subsubsection{Intervention} 

Using the data we had available from all channels, we finetuned MT5 \cite{xue2021mt5} for training our translation model in both the Hindi-Gondi and Gondi-Hindi direction. The model used in this work was adapted from \cite{https://doi.org/10.48550/arxiv.2210.15184} and we replicate the training setup specified there. The models were evaluated using sacrebleu (v2.2.0) with the spm tokenizer. As explored by \cite{wang2022expand}, we used the Hindi-Gondi dictionary to generate 200K noisy monolingual data to train our models, which we further used for continual pretraining. We also generated forward translated data using the same model to fuse with the lexicon-adapted data. Additionally, we utilize the quantized model developed by \cite{https://doi.org/10.48550/arxiv.2210.15184} to evaluate the performance of the same model's compressed version (400 MB in disk size compared to 2.28 GB), for the purpose of commenting on the edge deployment feasibility in low-resource environments.

The annotation setup is same as the one described in section \ref{sec:annotation-setup}. All annotators scored 40 inferences generated from the model to analyze its quality.

\begingroup
\setlength{\tabcolsep}{3pt} 
\renewcommand{\arraystretch}{1} 
\begin{table}[h]
    \centering
    \small
    \begin{tabular}{lccc}
    \toprule
    \textbf{Model} & \textbf{BLEU} & \textbf{Size} \\
    \midrule
    mt5-small (Hindi-Gondi)& 14.3 & 1.2GB \\
    mt5-small (Gondi-Hindi)& 30.9 & 1.2GB \\
    mt5-base (Hindi-Gondi) & 15.6 & 2.2GB \\
    mt5-small with continued pretraining & 14.9 & 1.2GB \\ 
    mt5-small with fused data (With 50K forward translated) & 12.6 & 1.2GB \\ 
    Quantized MT5 & 13.8 & 400 MB\\
  
    \bottomrule
    \end{tabular}
    \caption{Performance of the Hindi-Gondi translation model with model adaptations like lexicon-adaptation, continual pretraining and quantization.}
    \label{tab:model-scores}
\end{table}
\endgroup

\subsubsection{Takeaways}

As shown in Table \ref{tab:data-scores}, the annotators report an average DA Score of 63 for the Hindi-Gondi model which connotes that the model's output was semantically appropriate but contained an identifiable degree of grammatical errors and typos. The annotators mentioned that the model's output was visibly poor in terms of producing appropriate punctuation. While the structural integrity of the sentence seems to have been preserved, spelling differences and code-switched tokens ( i.e. Hindi tokens on the target side ) were common. Annotators also observed the occurrence of ambiguous-language tokens that they were unable to identify as being correct in the given context, which could be attributed to the dialectal differences that existed within the language. 

The BLEU scores of the models are reported in Table \ref{tab:model-scores}. The significant drop in the BLEU score in the Hindi-Gondi direction, in comparison to Gondi-Hindi, appears to be most significantly associated with the inclusion of Hindi in the pretraining corpus of MT5. The mC4 training corpus contains 1.21\% Hindi data, which greatly appears to aid the target side reconstruction in the Gondi-Hindi model \cite{xue2020mt5}. Additionally, we posit that the mt5 tokenizer, which has never seen Gondi, expectedly falters in the reconstruction of Gondi on the target side ( as is seen in the Hindi-Gondi ) model, so we expect a high degree of tokenization ambiguities in the Gondi output which would adversely impact the BLEU evaluation. This is further substantiated by observation of annotators who claimed that in some cases constructions of a Gondi token emulated the Hindi construction of the token. We intend to do a deeper investigation of such spurious correlations that massively-multilingual models might develop when a dominant attribute of language, its script in this case, matches other well-represented languages in the pretraining corpus despite being linguistically quite divergent. 

One of the primary learnings from this task was the need to integrate linguistically-motivated validation structures in a data collection channel to optimize the yield of the data collection process. In our specific case, the lack of any validation structures around the dialects against which we collected our data for led to the collection of a dataset with samples from at least 3 dialects. This especially harmed the impact of the machine translation system as it's responses catered to neither dialect completely.

\subsection{Assistive Translation}

We built an interface showing suggestions from our translation model to see if it improves the quality or speed of translation efforts.

\subsubsection{Motivation}

We use the same Hindi-Gondi model to generate candidate words that could be used when translating Hindi sentences to Gondi. The motivation for this task was to understand if poor-accuracy models could be used to accelerate the data collection process by providing assistive recommendations to the participating translators. This task holds unique practical value because it allows us to comment if sub-optimal machine translations (which would be an obvious intermediate artifact in such community interventions) can help enhance the yield of data-generation pipelines in low-resource communities. 

 \begin{figure}[h]
    \begin{center}
    \includegraphics[scale=0.15]{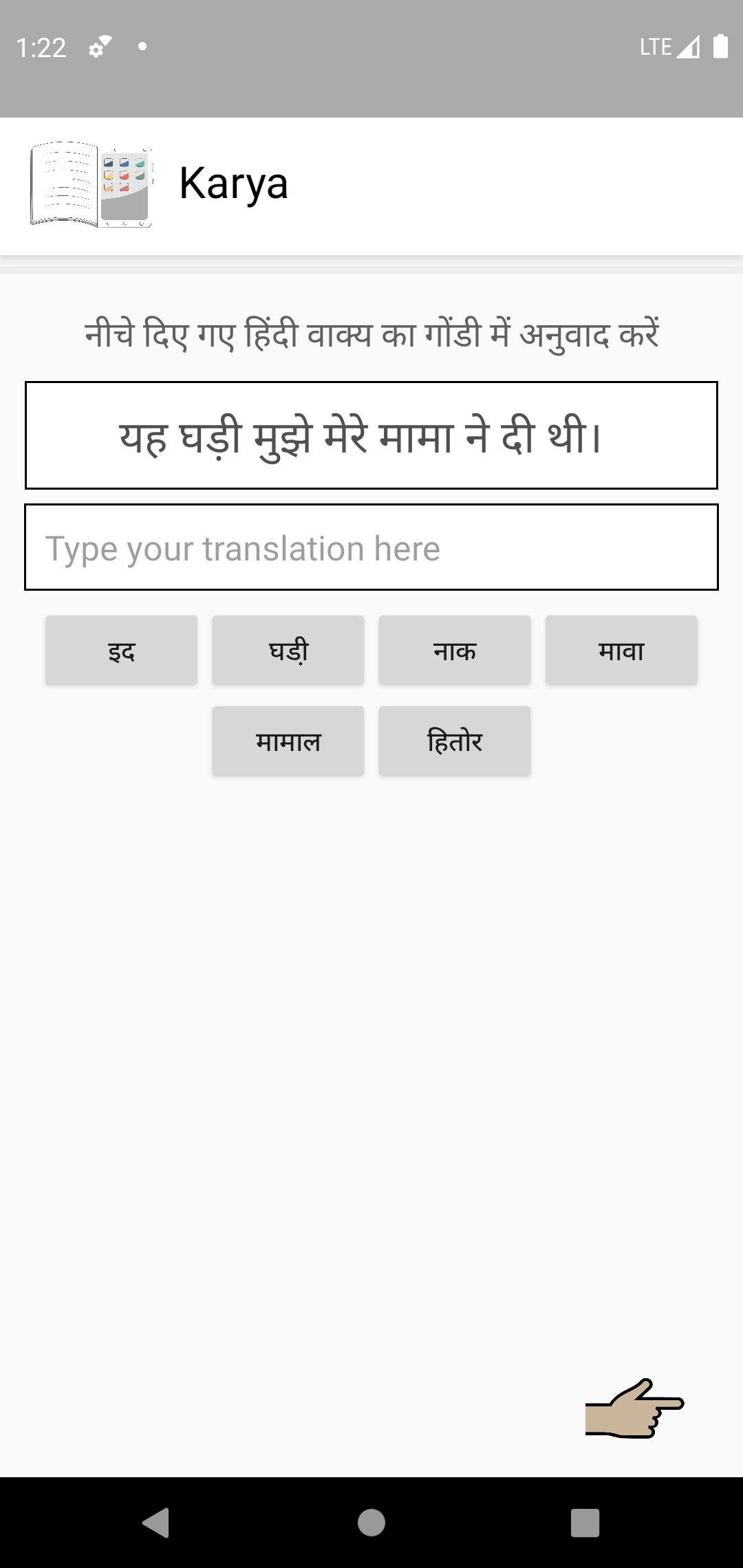} 
    
    \caption{Gisting Interface for providing recommendations to data-providers.}
    \label{fig:model-gisting}
    \end{center}
    \end{figure}

Our motivation to develop this system did not come directly from the Gond community but through partners at an educational institution and the citizen journalism platform CGNet Swara. They believed it could enable speedy translation of books for primary education and media stories into Gondi. 


\subsubsection{Intervention}

We realized that developing such a language technology had to take local constraints into account, which in this specific case entailed the lack of a stable internet connection. We thus developed an offline variant of the model, which was compressed enough to operate on relatively low-resource devices without requiring an internet connection. 

The interface for this task, as shown in Figure\ref{fig:model-gisting}, is a screen that shows users the model's output as a set of candidate tokens that they can opt for while submitting the translation. We sample 150 sentences from our development set and provided it for translation through the Karya app. Inferences for these sample sentences were provided to the annotators as a panel of assistive candidate tokens. The annotators were requested to post-edit these candidate gists until they were correct using the provided suggestions wherever possible. During this exercise we monitor the number of suggestions accepted by the annotators over the total number of suggestions presented. 

\subsubsection{Takeaways}

We find that annotators use 3.66 out of an average of 5.56 options shown to them in each sentence translation iteration. This is further substantiated by the observations of the annotators, who reported that nearly half of the suggestions provided by the model were accurate and useful in generating the final output. They also commented on the interface design by mentioning that they preferred to type a token displayed in the panel of candidate tokens instead of clicking on the token to avoid disrupting their flow while typing out a translation. This indicates that tracking the number of accepted tokens via clicks would have provided us a pessimistic view of the model's usefulness since annotators were visually aided by tokens (which might have been completely or partially correct) and our experimental design would not have logged that as a successful instance of aid. Annotators also pointed out that some stop words were useful candidates in the Bag Of Words interface as they preferred opting for them while typing a translation (rather than post-editing them as a part of a gist). Hence, providing even noisy assistance with the right interface appears to be a promising direction. 

The annotators overarching evaluation of the translation system deemed it to be semantically accurate. This gives hope that we can setup a virtuous cycle where generation of linguistic resources enables the development of community-oriented language technologies, which in turn could help in generating more linguistic resources that can then further improve language technologies. Additionally, their feedback on finding the candidate translations being visually assistive encourages a deeper investigation into if and how relatively low-accuracy models could be leveraged to improve the experience of data providers in low-resource communities.



\subsection{Disseminating Gondi content via Interactive Voice Response}

Collecting data and building language technologies is only one part of the equation; the other is creating avenues for communities to hear their language broadcast over mass media and other communication channels.

\subsubsection{Motivation} There is an established body of work on the role of Interactive Voice Response (IVR) forums for reaching communities that are too poor to afford smartphones, too remote to access the Internet or too low-literate to navigate text-driven environments \cite{swaminathan2019learn2earn,dipanjan2019,revisitCG,designLess,ivrFarmers,raza2018baang,sawaal,raza2013job,sangeet}. As an example, Learn2Earn \cite{swaminathan2019learn2earn} is an IVR based system that was used as an awareness campaign tool to spread farmers' land rights in rural India, HIV literacy and voter awareness \cite{mehta2020using}. Learn2Earn awards mobile talktime to users who call a toll-free number and answer all multiple-choice questions on the message correctly. Starting from an initial set of just 17 users, it was successful in spreading land rights awareness to 17,000 farmers in 45 days via additional rewards for successful referrals \cite{swaminathan2019learn2earn}.

We adapted Learn2Earn for spreading voter awareness among Gondi speakers in Dantewada (a rural district in the state of Chhattisgarh in India), during the time that a bypoll election was being held. Prior work has shown that larger the contexts, identities and communicative functions associated with a language, the more likely it is to thrive \cite{walsh2005will}. Voter rights content in local languages during election season has the potential to encourage conversations on topics of wider contexts and functions as well as contribute to establishing representative and effective governance.

\subsubsection{Intervention} 

\begin{figure*}[!t]
    \centering
    \includegraphics[width=\linewidth,keepaspectratio]{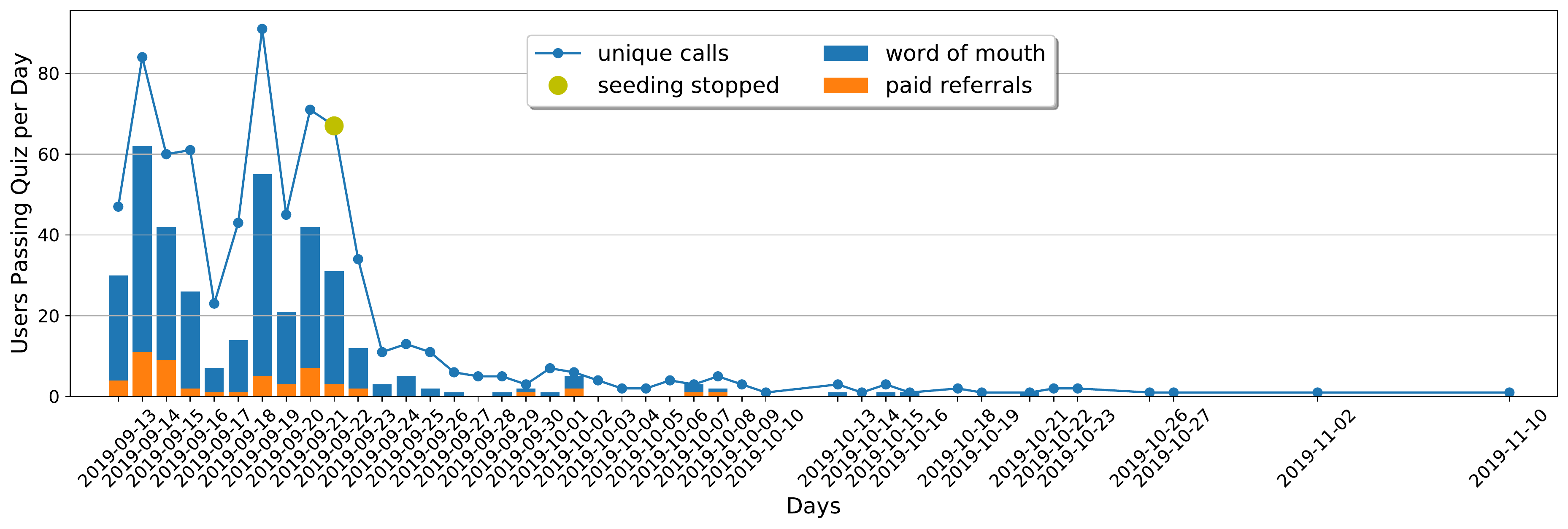}
    \caption{Number of calls to the system (blue line) and the number of users who answered all questions correctly (bar plot). The blue and orange bars represent users onboarded via seeding and through referrals}
    \label{fig:usage}
\end{figure*}

The pilot obtained users through seeding activities and referrals. Figure\ref{fig:usage} shows the number of unique calls that were made to our system and the number of users who passed the quiz over time. Further, the Figureshows that a majority of quiz passers came to know about our system via direct seeding, though people knew that additional credits can be earned through referrals. Usage dropped sharply after the elections in September concluded and seeding activities were discontinued. 

\definecolor{lightgray}{gray}{0.9}
\begin{table*}
\small
\rowcolors{2}{lightgray}{}
{\renewcommand{\arraystretch}{\pad}
\begin{tabular}{! \vbl p{3.5in}|l! \vbl}
\hbl
\cellcolor[HTML]{C0C0C0} \textbf{Metrics}                                                                                   & \cellcolor[HTML]{C0C0C0} \textbf{Value}     \\ \hline
\begin{tabular}[c]{@{}l@{}}Unique callers (total, during and after seeding)\end{tabular}     & (557, 480, 77)   \\ \hline
\begin{tabular}[c]{@{}l@{}}Unique callers per day - during seeding (min, mean, median, max)\end{tabular}     & (13, 48, 49, 80)   \\ \hline
\begin{tabular}[c]{@{}l@{}}Unique callers per day - after seeding (min, mean, median, max)\end{tabular}     & (0, 4, 3, 20)   \\\hline
\begin{tabular}[c]{@{}l@{}}Callers answering all questions correctly\end{tabular} & 313 \\ \hline
\begin{tabular}[c]{@{}l@{}}Callers answering all questions correctly in their first call\end{tabular} & 104               \\ \hline

\begin{tabular}[c]{@{}l@{}}Calls made by callers (min, mean, median, max)\end{tabular}          & (1, 3, 2, 64)      \\ \hbl
\end{tabular}}
 \caption{Summary statistics for our Learn2Earn deployment in Gondi}
\label{tab:summary}
\end{table*}

Table \ref{tab:summary} highlights some important statistics in our Learn2Earn pilot. We spread voter awareness in Gondi to 557 speakers, of which a majority (86\%) were reached during active seeding. Nearly 60\% of the users correctly answered all questions, either in their first attempt or in successive attempts, indicating the effectiveness of the system for content dissemination. Further, many users called the system more than once, with one user calling 64 times (to refer someone, users had to place another call to the system). A follow-up survey revealed that 22 of 113 users surveyed could not vote as they did not have a voter ID card, potentially useful information for authorities to increase voter turnout. One disappointing finding was that only about 8.3 percent of users were women.

\subsubsection{Takeaways} We see three clear benefits from conducting Learn2Earn pilots in endangered or vulnerable languages. First, since it is entirely in the spoken form, only native speakers of an endangered or vulnerable language can comprehend the content and earn the reward. It is also easily accessible in low-resource environments as it requires a missed call and nothing more. Second, the phone numbers collected are an important dataset of speakers of that language and can be used for future translation workshops and related programs that help their economic and linguistic development. Finally, an oral language has a tendency to die out unless there is an opportunity for that language to be used outside of everyday surroundings, which periodic Learn2Earn campaigns on important issues can help achieve.

\section{Discussion}

Steven Bird alerts us to the dangers inherent in a technology driven approach to language revival, contending that commodifying indigenous languages as data alienates native speakers. He argues for an approach rooted in self-determination where outside experts are only engaged with to help implement a program or strategy that the community has already settled on \cite{bird2020decolonising}. This sentiment is echoed by Stebbins et al (2017) who maintain that `the community must be the driver and driving interest of the research and language projects"\cite{stebbins2017living}. Lewis and Simons carve out an additional role for reflective practitioners of language development, that of "facilitating opportunities for local language users to interact with each other" \cite{book}.

There are also concerns that experts profit off their domain expertise in an indigenous language with little benefit accruing to the communities themselves. For example, Bird asks "where are the full implementations, deployed in robust software, in active use to capture primary data, leading to curated language products that are being mobilized in speech communities?" By his yardstick, even the interventions described in our paper fall short as none have translated to scale or persistent use by the community. With the benefit of hindsight, we can now reflect on whether our initiatives have been exploitative. 

Our position is that these interventions avoided an exploitative approach by providing tangible outputs advancing one of the five FAMED conditions laid out by Lewis and Simons as an ethical guide to language practitioners. This acronym expands to having Functions associated with the language that are in use and recognized by the community; a means for community members to Acquire proficiency in the language; a Motivation for community members to use the language, which is frequently but not exclusively economic; a policy Environment that is conducive to the growth of the language; and a Differentiated sphere for use of the language in adherence to established societal norms. 

Our machine translation artifact can be built upon by the Gondi community and other low resource communities around the world to expand the \textit{functions} associated with their language, through automated translation of digital content. The assistive translation app can help in language \textit{acquisition} and documentation by improving the speed of translating literary material and providing vocabulary that translators may themselves be unaware of. Our Learn2Earn pilot provided incentive payments to those speaking Gondi, thus increasing (at a small scale) the \textit{motivation} to use and recognize the language. Translating children's stories into Gondi can create a fertile policy \textit{environment} that makes possible the use of tribal languages in primary education. The dictionary development process kickstarted a \textit{differentiated} sphere of Gondi usage between tribals of different states, whereas it has historically been used only within their own region. Our thematic approach of collecting language data to build technologies while simultaneously creating resources for the community that advance one of the FAMED conditions can help avoid the pitfall of alienating local speakers by commodifying their language as data.

Overall, our work raises interesting questions on the role of the outside interventionist in building langugage technologies and the relationship they should have with the community. For example, the dictionary development workshops primarily comprised bilingual and educated Gond speakers, which may have contributed to their desire for standardizing the language. However, we realized that this approach would be discriminatory to monolingual Gond speakers who would have to accept the language thrown at them by the intellectuals within their community since the standardization efforts would not be universally agreed upon or even known to the majority of Gond speakers. As Bird writes, "No community speaks with a single voice, and in building relationships we can unwittingly align ourselves with agendas, clans and gate-keepers." This prompts reflection on whether the outside interventionist should simply fill in "white man's paperwork" and let the community lead, or if there is a deeper role that we need to play, one where we use our perceived neutrality to arbitrate between the different sections of a community. The latter role finds expression in Lewis and Simons' contention that "it may be an appropriate role for an outsider to act as an intermediary between the different speech communities." While we do not presume to have answers to these questions, we hope that our work helps stimulate wider debate in the language development community on defining our relationship with the speech communities we work with and depend upon.

\section{Conclusion}

Engineering is the primary obstacle to designing technologies for well-resourced languages. For low-resource ones, more focus needs to be given towards designing methods for robust data collection and evaluation upon which the language technology can be built.

To keep community members motivated through the data collection process, our team strived to achieve 2 simultaneous goals: the collection of data upon which language technologies such as speech to text or machine translation could be built, and building a literary resource that community members could point to as an immediate, demonstrable success. For example, the Learn2Earn pilot in Gondi not only provided the community with an opportunity to earn money for answering a quiz in their language and referring others to it, but it also provided a dataset of native Gondi speakers. Similarly, translating children's stories and creating a standardized dictionary resulted in both data upon which a machine translation tool was built and also tangible language resources that can be used by the community.

We made use of the data collected to develop a machine translation model and an assistive translation interface. Community evaluators reported that the model fared poorly in punctuation and grammar, and roughly half the suggested words were applicable to the sentence they were translating. Our future goals are improving the models accuracy by feeding it more data and testing its performance with data from other low-resource languages.

In 2020, the Chhattisgarh government began allowing primary education in 10 tribal languages, including Gondi \footnote{https://indianexpress.com/article/governance/chhattisgarh-education-reforms-tribal-languages-to-be-a-medium-of-education-in-pre-school-6271547/}. At the community level, our focus is on working with the government to integrate the linguistic resources and technological artifacts we've created into their apparatus. We also plan on integrating the translation tool with CGNet Swara's citizen journalism newsroom so that they can regularly put out Gondi content.

\bibliographystyle{spmpsci}      
\bibliography{refs}

\end{document}